\def\year{2018}\relax
\documentclass[letterpaper]{article} 
\usepackage{aaai18}  
\usepackage{times}  
\usepackage{helvet}  
\usepackage{courier}  
\usepackage{url}  
\usepackage{graphicx}  
\usepackage{latexsym}
\usepackage{amsmath}
\usepackage{amssymb}
\usepackage{algorithm}
\usepackage{algpseudocode}
\usepackage{caption}
\usepackage{subcaption}
\usepackage{graphicx}
\usepackage{booktabs}
\newcommand{\citet}[1]{\citeauthor{#1}~\shortcite{#1}}

\usepackage{hhline}
\usepackage{pgf}
\usepackage[export]{adjustbox}
\usepackage{amsfonts}      
\usepackage{graphicx}
\usepackage{amssymb}%

\begin{document}
%
\title{Using Syntax to Ground Referring Expressions in Natural Images}

\author{
  Volkan Cirik, Taylor Berg-Kirkpatrick, Louis-Philippe Morency \\
School of Computer Science\\
Carnegie Mellon University\\
Pittsburgh, PA 15213 \\
\texttt{\{vcirik,tberg,morency\}@cs.cmu.edu} \\
}

\maketitle

\begin{abstract}
We introduce GroundNet, a neural network for referring expression recognition -- the task of localizing (or grounding) in an image the object referred to by a natural language expression.
Our approach to this task is the first to rely on a syntactic analysis of the input referring expression in order to inform the structure of the computation graph.
Given a parse tree for an input expression, we explicitly map the syntactic constituents and relationships present in the tree to a composed graph of neural modules that defines our architecture for performing localization. 
This syntax-based approach aids localization of \textit{both} the target object and auxiliary supporting objects mentioned in the expression.
As a result, GroundNet is more interpretable than previous methods: we can (1) determine which phrase of the referring expression points to which object in the image and (2) track how the localization of the target object is determined by the network.
We study this property empirically by introducing a new set of annotations on the GoogleRef dataset to evaluate localization of supporting objects.
Our experiments show that GroundNet achieves state-of-the-art accuracy in identifying supporting objects, while maintaining comparable performance in the localization of target objects.
\end{abstract}
\newcommand \figurenine{
\begin{figure*}[t]
\begin{subfigure}{.50\textwidth}
\centering
\includegraphics[width=0.60\linewidth]{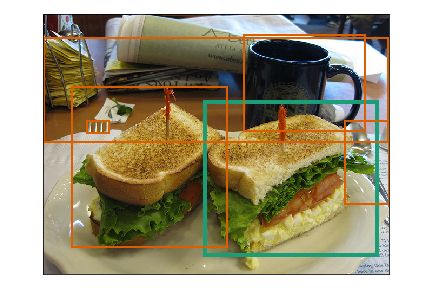}
\caption{An example referring expression from our validation set \textit{``half of a sandwich on the right side of a plate nearest a coffee mug''}.
Orange boxes are region candidates and green box is the referred bounding box.}\label{fig1:a}
\end{subfigure}%
\begin{subfigure}{.50\textwidth}
\centering
\includegraphics[width=0.90\linewidth]{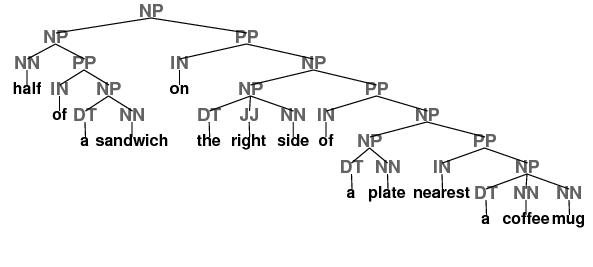}
\caption{The parse tree for the referring expression in (a).}\label{fig1:b}
\end{subfigure}%
\vfill
\begin{subfigure}{.49\textwidth}
\centering
\includegraphics[width=0.99\linewidth]{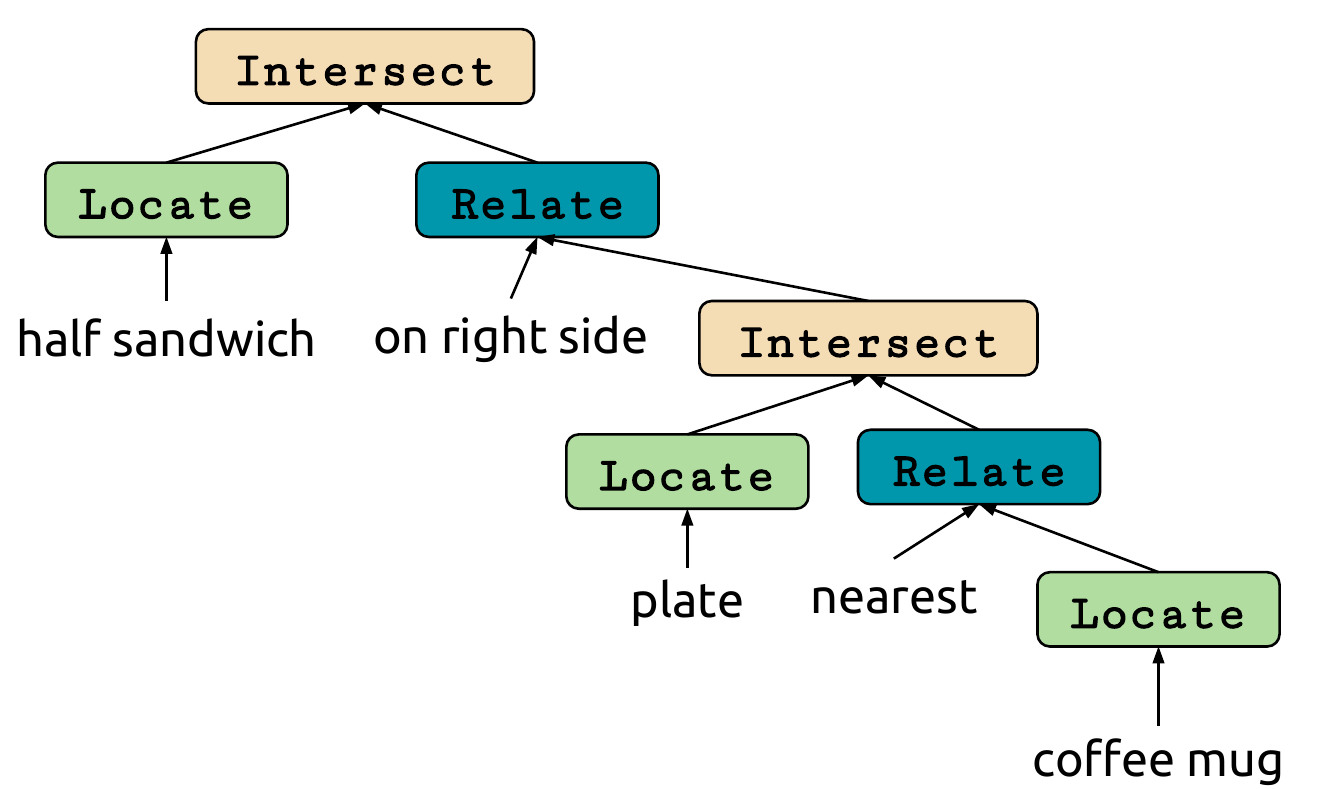}
\caption{Computation graph for the parse tree in (b).}\label{fig1:c}
\end{subfigure}
\begin{subfigure}{.49\textwidth}
\centering
\includegraphics[width=0.85\linewidth]{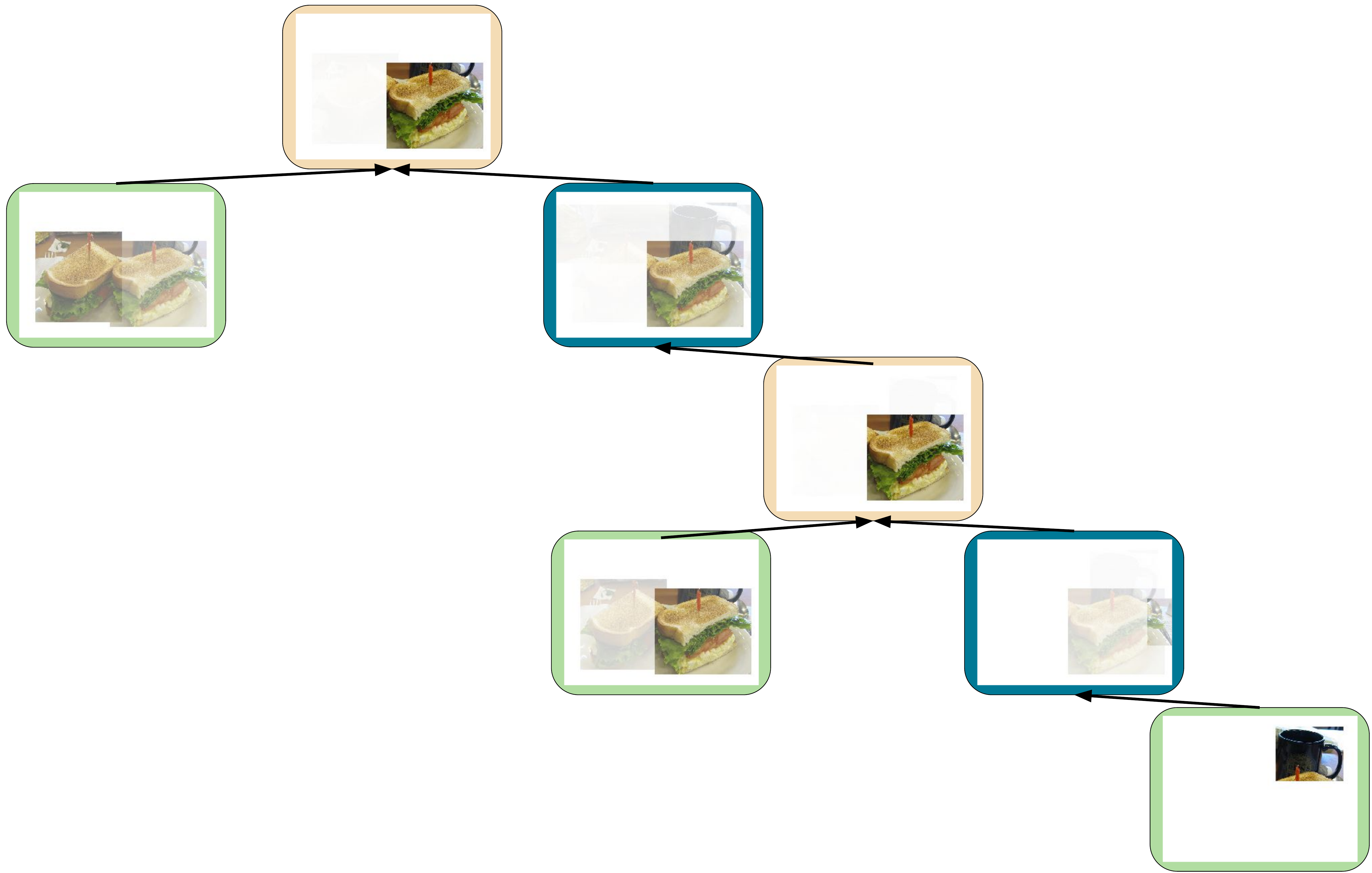}
\caption{Grounding of objects in (a) with the computation graph in (c). The more visible objects have higher probabilities. Note that the model is able to ground supporting objects like the coffee mug.}\label{fig1:d}
\end{subfigure}
\caption{An Overview of  GroundNet.  A referring expression (a) is first parsed (b). Then, the computation graph of neural modules is generated using the parse tree (c). Each node localizes objects present in the image  (d).}\label{fig1}
\end{figure*}
}

\newcommand \figureten{
\begin{figure*}[t]
\centering
\includegraphics[width=0.85\linewidth]{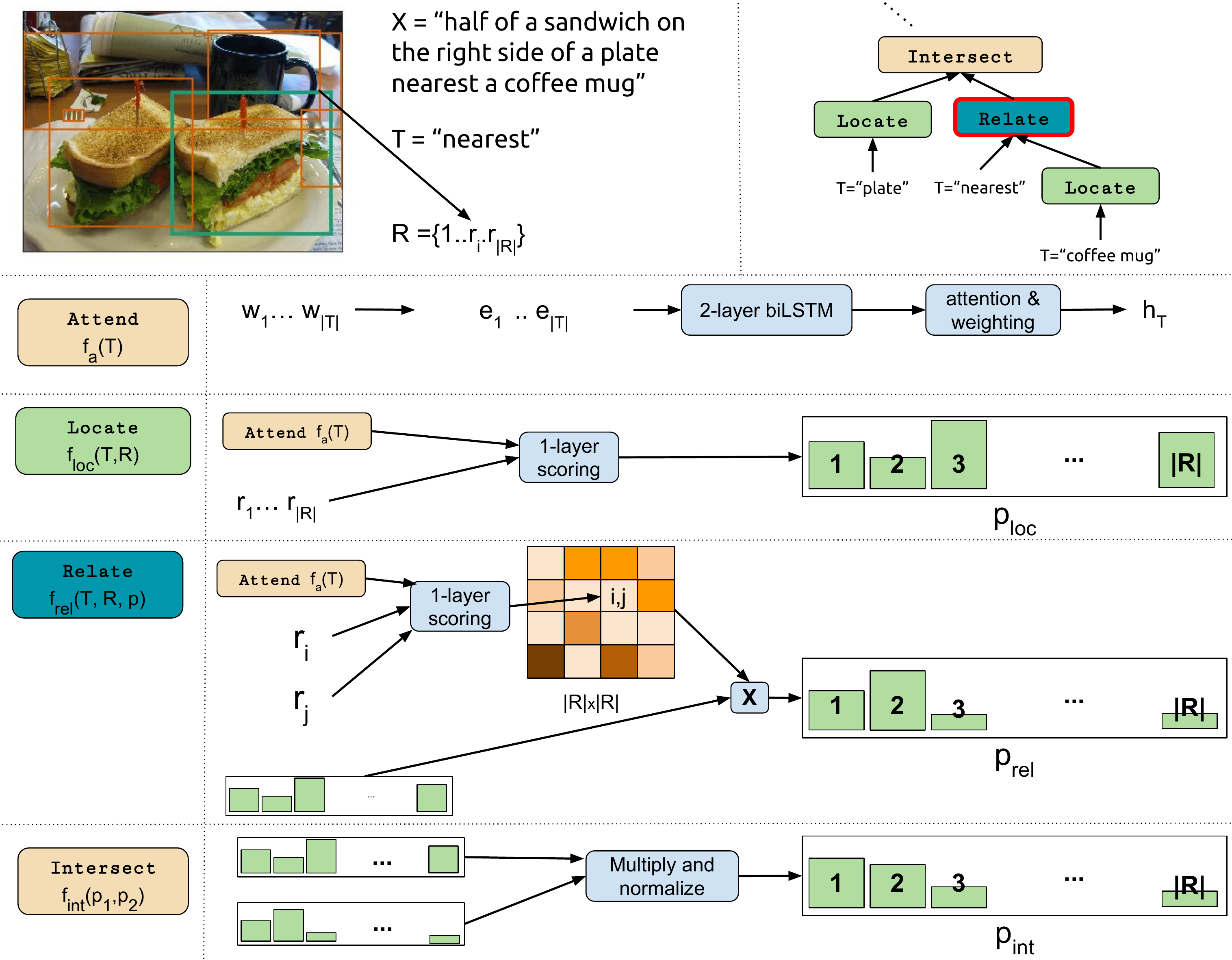}
\caption{Illustrations of GroundNet's neural modules.
Upper left shows an example referring expression and the input for \texttt{Relate} node (upper right, highlighted in red) of a small section of a computation graph.
Modules take inputs from module's text span $T$, the set of bounding boxes $R$, and output probabilities of other nodes $p_i$.
Best seen in color.}\label{fig2}
\end{figure*}
}

\newcommand \tableseven{
\begin{table*}[t]
\centering
\small
\begin{tabular}{@{}lcccccr@{}}
\toprule
Model&Syntax&Dynamic Computation&Modularity&Relationships&Supporting(\%)&Accuracy(\%)\\ \midrule
LSTM+CNN - MMI &  &  &  &  &  & 60.7 \\
LSTM+CNN - MMI+visdif &  &  &  & $\checkmark$ &  & 64.0 \\
LSTM+CNN - MIL &  &  &  & $\checkmark$ & 15.0 & 67.3 \\
CMN  &  &  & $\checkmark$ & $\checkmark$ & 11.1 & 69.7 \\ \midrule
Recursive NN & $\checkmark$ & $\checkmark$ &  &  &  & 51.5 \\
CMN-syntax guided  & $\checkmark$ &  & $\checkmark$ & $\checkmark$ &  & 53.5 \\ \midrule
GroundNet & $\checkmark$ & $\checkmark$ & $\checkmark$ & $\checkmark$ & 60.6 & 65.7 \\
GroundNet-syntax-guided \texttt{Locate}& $\checkmark$ & $\checkmark$ & $\checkmark$ & $\checkmark$ & 60.0 & 66.7 \\
GroundNet-free-form & $\checkmark$ & $\checkmark$ & $\checkmark$ & $\checkmark$ & 10.6 & 68.9 \\ \bottomrule
\end{tabular}
\caption{The accuracy of models with the support of syntax, dynamic computation, modularity, relationship modeling, and supporting object predictions. Our model is the first syntax-based model with successful results and achieves the best results in supporting object localization.}\label{tab1}
\end{table*}
}
\newcommand \figurecpick{
\begin{figure*}
\centering
\begin{subfigure}{.93\textwidth}
\centering
\includegraphics[width=0.99\linewidth]{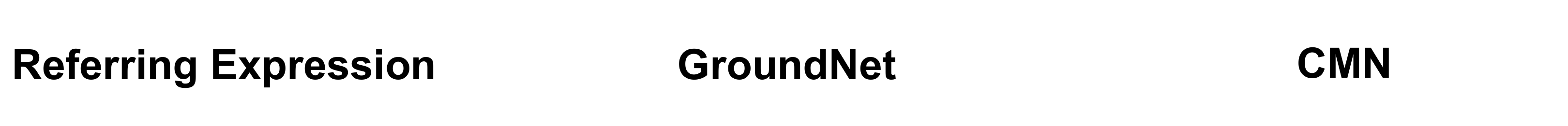}
\end{subfigure}%
\vfill
\centering
\begin{subfigure}{.31\textwidth}
\centering
\includegraphics[width=0.99\linewidth]{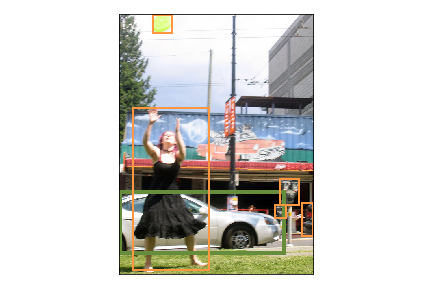}
\caption*{``a white color car behind a girl catching a disc''}
\end{subfigure}%
\centering
\begin{subfigure}{.40\textwidth}
\centering
\includegraphics[width=0.99\linewidth]{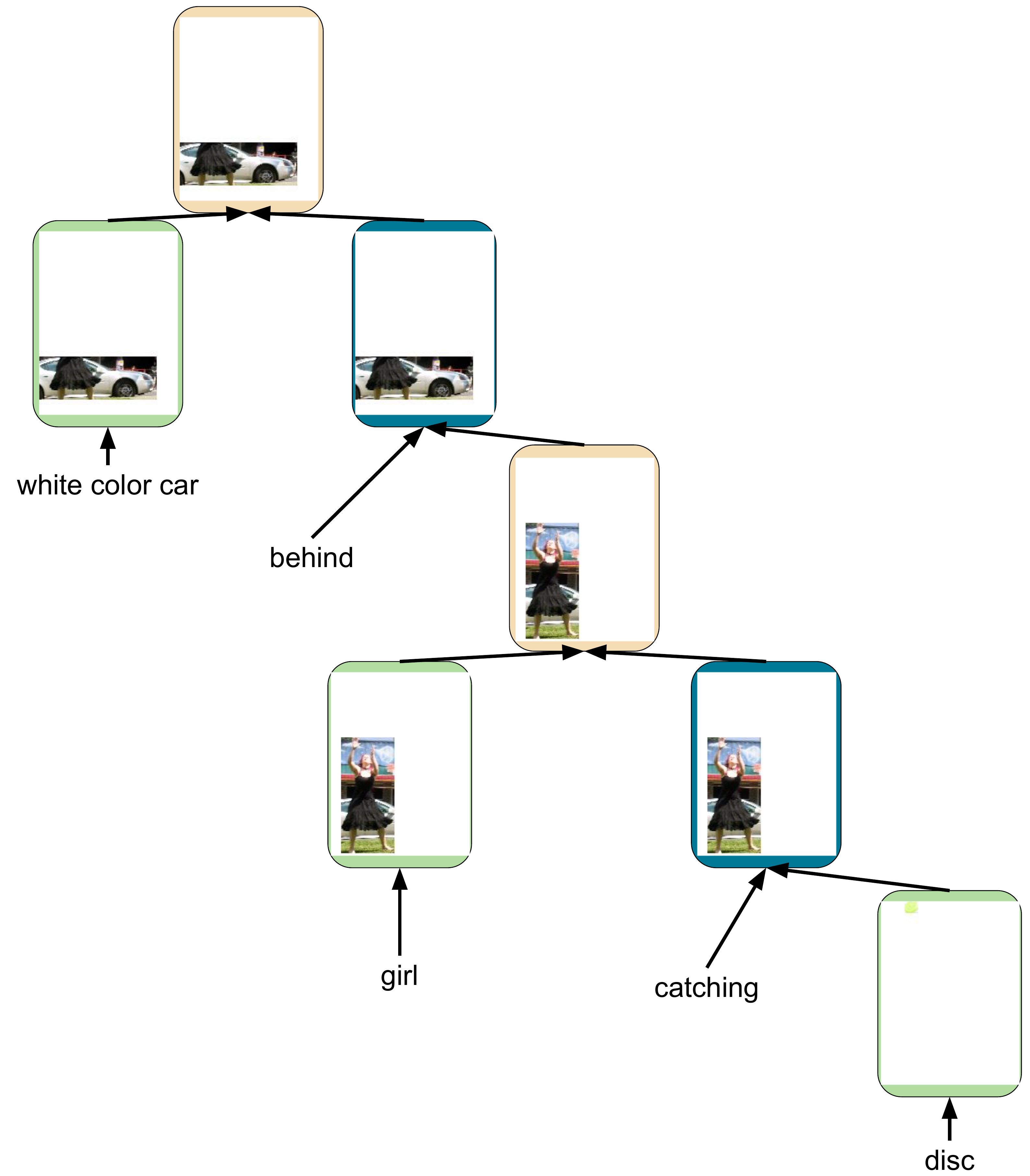}
\end{subfigure}%
\begin{subfigure}{.20\textwidth}
\centering
\includegraphics[width=0.99\linewidth]{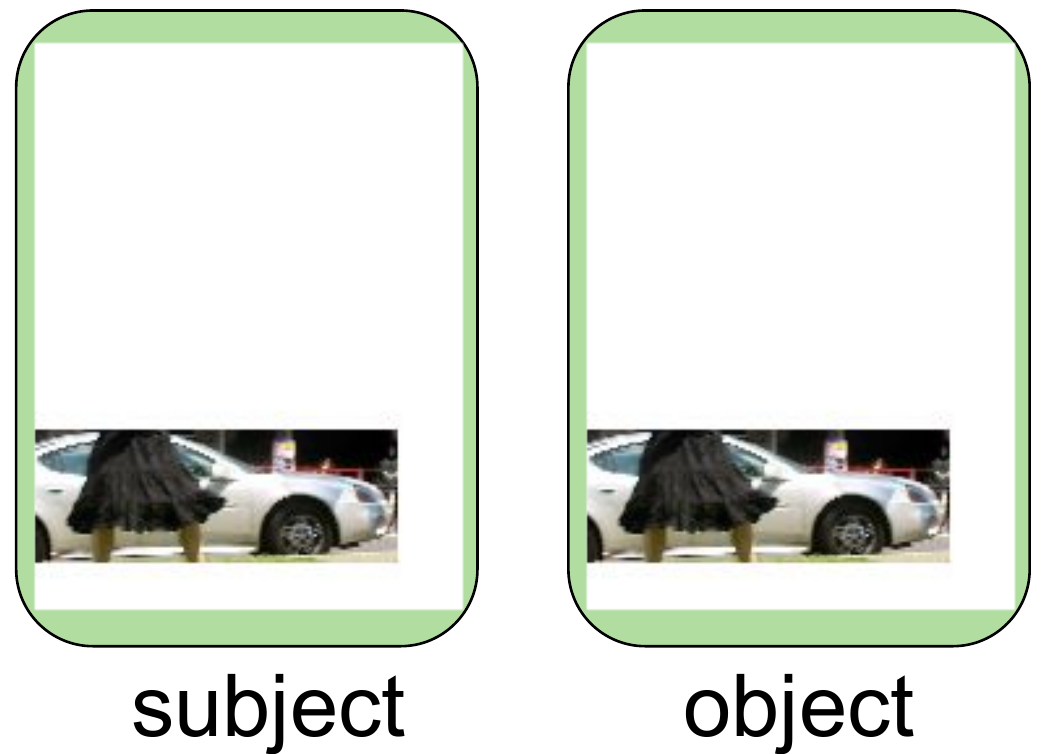}
\end{subfigure}%
\vfill
\centering
\begin{subfigure}{.31\textwidth}
\centering
\includegraphics[width=0.99\linewidth]{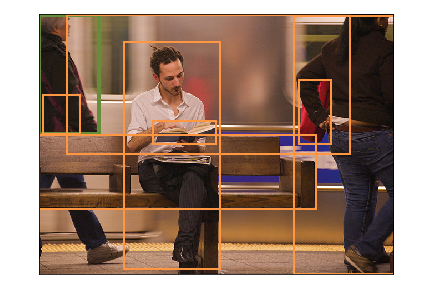}
\caption*{``the man walking behind the bench''}
\end{subfigure}%
\centering
\begin{subfigure}{.40\textwidth}
\centering
\includegraphics[width=0.99\linewidth]{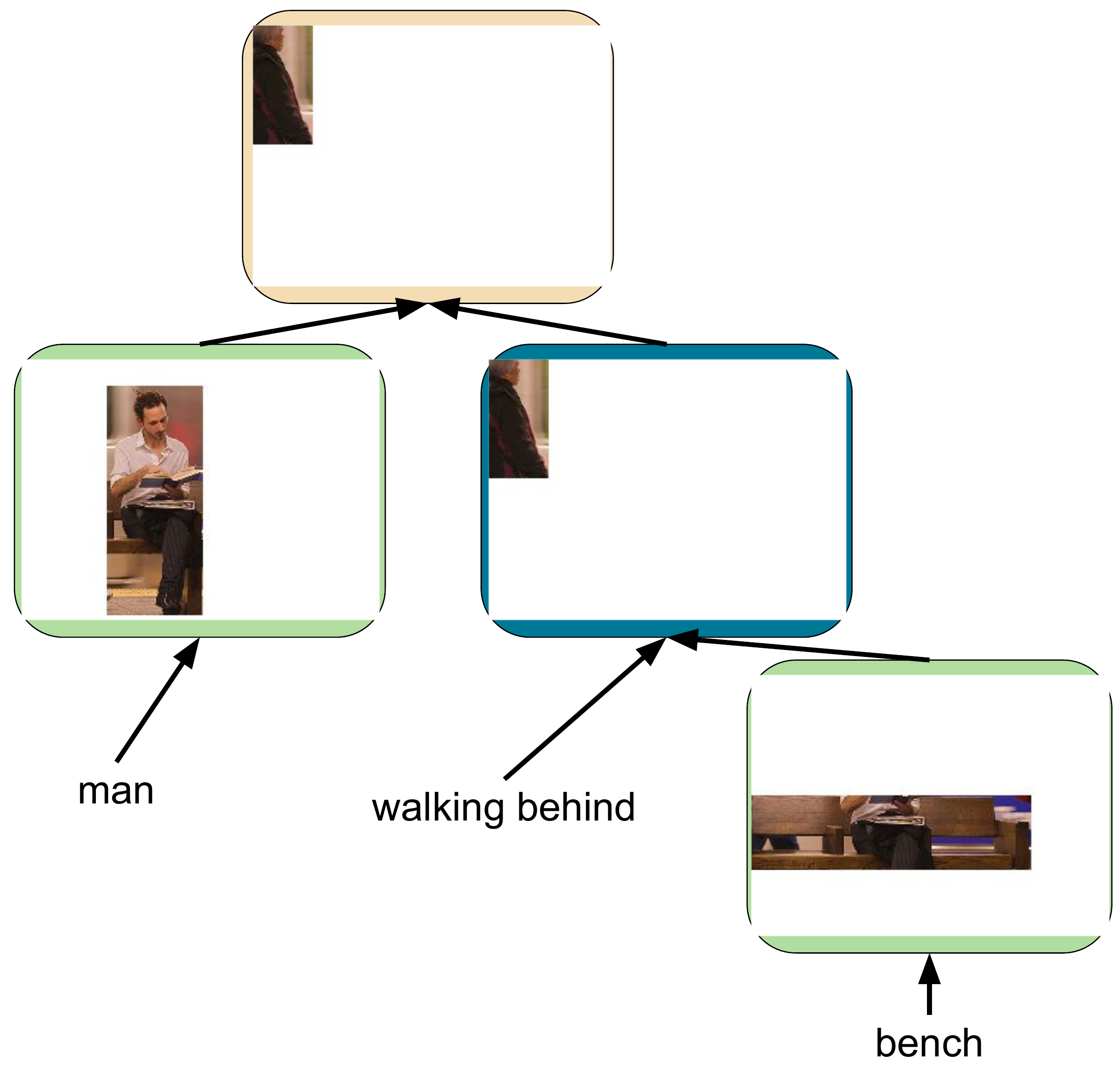}
\end{subfigure}%
\centering
\begin{subfigure}{.20\textwidth}
\centering
\includegraphics[width=0.99\linewidth]{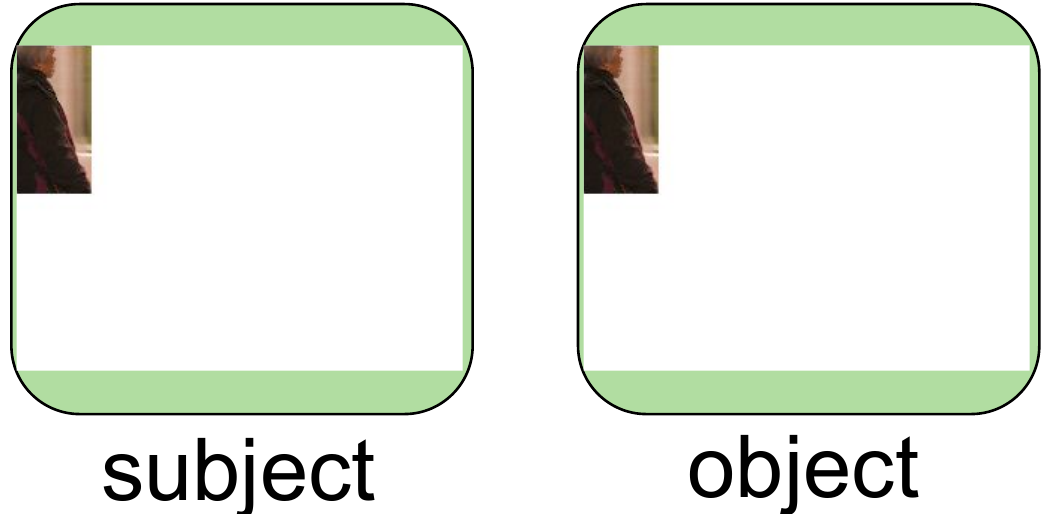}
\end{subfigure}
\vfill
\centering
\begin{subfigure}{.31\textwidth}
\centering
\includegraphics[width=0.99\linewidth]{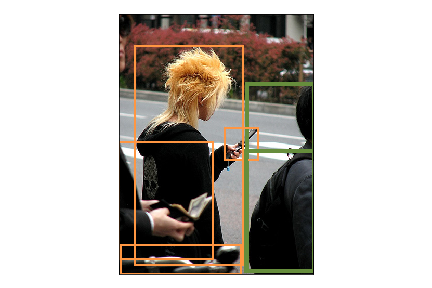}
\caption*{``a man going before a lady carrying a cellphone''}
\end{subfigure}%
\centering
\begin{subfigure}{.35\textwidth}
\centering
\includegraphics[width=0.99\linewidth]{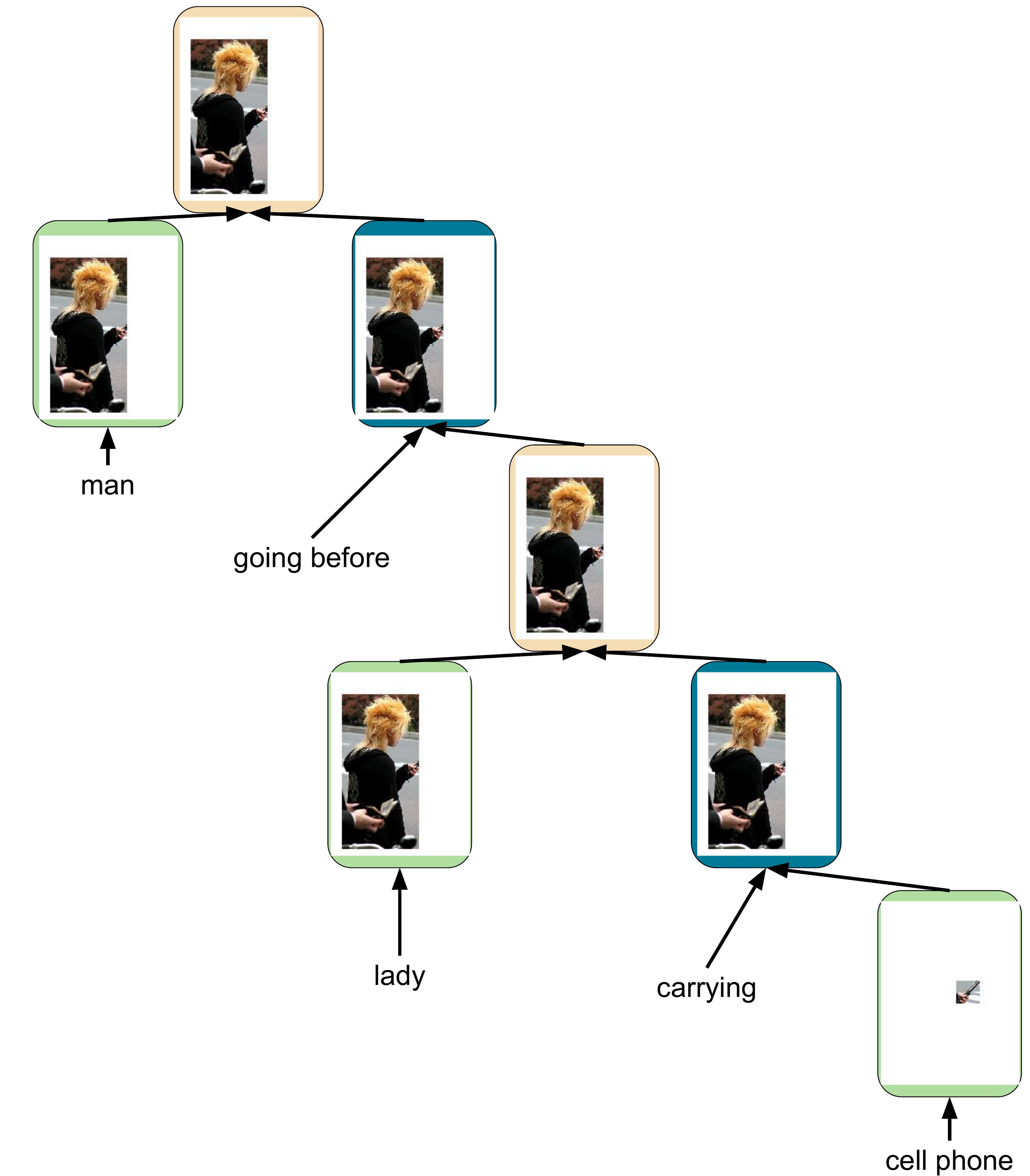}
\end{subfigure}%
\centering
\begin{subfigure}{.20\textwidth}
\centering
\includegraphics[width=0.99\linewidth]{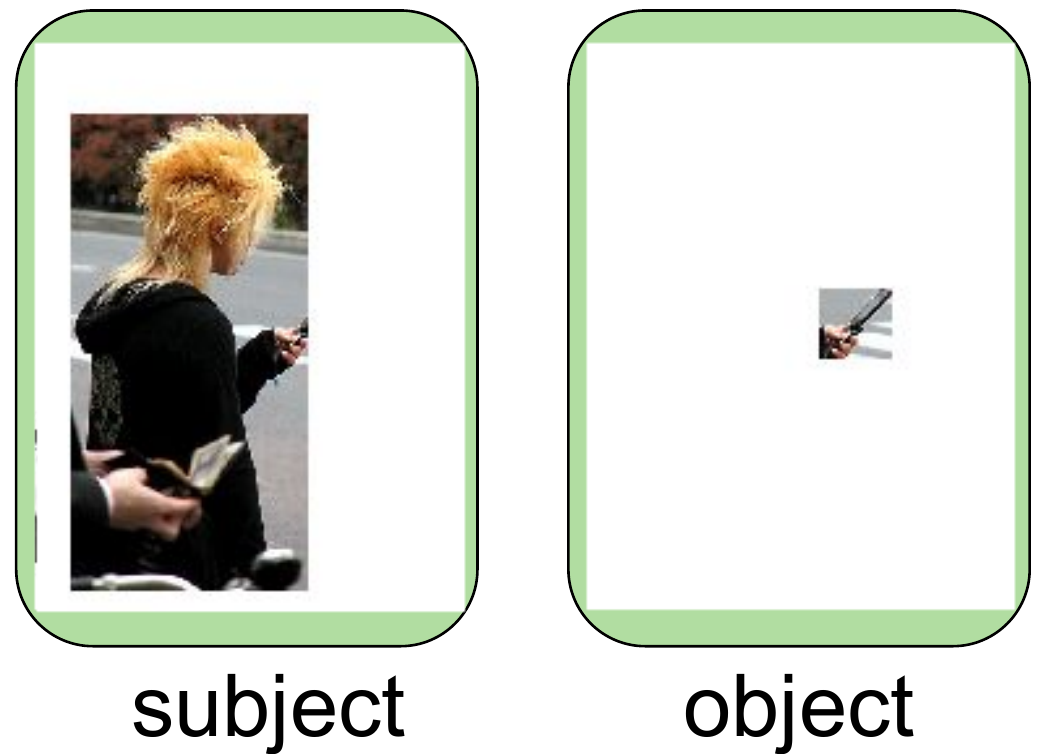}
\end{subfigure}%
\caption{Qualitative Results for GroundNet. Bounding boxes and referring expressions to target object (in green boxes) on the left.
GroundNet predictions in the middle and CMN
predictions are on the right.
GroundNet localizes not only the target object but also supporting objects (e.g. disc and girl in the first row, bench in the second).Best seen in color.} \label{fig3}
\end{figure*}
}
\section{Introduction}\label{sec:introduction}
Spatial referring expressions are part of our everyday social life (``Please drop me at the blue house next to the red mailbox.'') and also part of professional interactions (``Could you pass the small scalpel to the right of the forceps?'').
These natural language expressions are designed to uniquely locate an object in the visual world.
The process of grounding referring expressions into visual scenes involves many intermediate challenges.
As a first step, we want to locate all the objects mentioned in the expression.
While one of these mentions refers to the target object, the other mentions (i.e. supporting object mentions) are also important because they were included by the author of the referring expression in order to disambiguate the target. In fact, \citet{grice1975logic} argued that supporting objects will only be mentioned when they are \emph{necessary} for disambiguation.
As a second step, we want to identify the spatial relationships between these objects.
Is the target to the left of the supporting object? Is it beneath it?
To make effective use of an identified supporting object, we must understand how this object is related to the target.
And finally, for many natural referring expressions, the process is recursive: a supporting object may itself be identified by a relationship with another supporting object.
As a result, models that reason about referring expressions must respect this hierarchy, processing sub-expressions before attacking larger expressions.
Modeling this compositionality is critical to designing recognition systems that behave in an interpretable way and can justify their decisions.

\figurenine

In this paper, we introduce GroundNet, the first dynamic neural architecture for referring expression recognition that takes full advantage of syntactic compositionality.
Past approaches, such as the Compositional Modular Networks (CMN) model \cite{hu2017modeling}, have relied on limited syntactic information in processing referring expressions -- for example, CMN tracks a single supporting object -- but have not modeled linguistic recursion and therefore is incapable of tracking multiple supporting objects.
As shown in Figure 1, our GroundNet framework relies on a syntactic parse of the input referring expression to dynamically create a computation graph that reflects the recursive hierarchy of the input expression.
As a result, our approach tracks intermediate localization decisions of all supporting objects. Following the approach of \cite{andreas2016neural,andreas2016learning}, this computation graph is translated into a neural architecture that keeps interpretable information at each step of the way, as can be seen in Figure~\ref{fig1:d}.

We additionally present a new set of annotations that specify the correct locations of supporting objects in a portion of the standard benchmark dataset, GoogleRef \cite{mao2016generation} to evaluate the interpretability of models for referring expression recognition.
Using these additional annotations, our empirical evaluations demonstrate that GoundNet substantially outperforms the state-of-the-art at intermediate predictions of the supporting objects, yet maintains comparable accuracy at target object localization.
These results demonstrate that syntactic compositionality can be successfully used to improve interpretability in neural models of language and vision. Our annotations for supporting objects and implementations are available for public use\footnote{\url{https://github.com/volkancirik/groundnet}}.
\section{GroundNet}\label{sec:method}

In this section, we explain the motivation of GroundNet, how we generate the computation graph for GroundNet, and finally, detail the neural modules that we use for computing the localization the referring expressions. 

\subsection{Motivation} 
A referring expression disambiguates a target object using the object's discriminative features such as color, size, texture etc., and their relative position to other \textit{supporting} objects.
Figure~\ref{fig1:a} shows a canonical example from our task w one half of a sandwich is referred by ``half of a sandwich on the right side of a plate nearest a coffee mug''.
Here the sandwich is disambiguated using relative clauses (e.g. ``the right side of'' , ``nearest'') and the \textit{supporting} objects (e.g ``plate'', ``coffee mug'').
We observe that there is a correspondence between the linguistic compositional structure (i.e. the parse tree) of the referring expression and the process of resolving a referring expression.
In Figure~\ref{fig1:b}, we see that the target object and supporting objects have a noun phrase (NP) on the parse tree of the referring expression.
Also, the relative positioning of objects in the image (e.g. being on the right, or near) correspond to prepositional phrases (PP) on the tree.
We design GroundNet based on this observation to localize the target object by modeling the compositional nature of the language.
The compositionality principle states that the meaning of a constituent is a function of (i) its building blocks and (ii) the recursive rules to combine them.
In our case, the building blocks for the GroundNet is grounding of objects i.e. the probability of how likely an object is for word phrases.
The combining rules are defined by the parse tree describing what these objects are and how they are related to each other. 

GroundNet models the processing of a referring expression in a computation graph (see Figure~\ref{fig1:c}) based on the parse tree of the referring expression (see Figure~\ref{fig1:b}).
Nodes of the computation graph have 3 different types aiming to capture the necessary computations for localizing the target object.
\texttt{Locate} nodes ground a noun phrase (``half sandwich'', ``plate'',``coffee mug''), i.e. pointing how likely that a given noun phrase refers to an object present in the image.
For example, in Figure~\ref{fig1:d}, \texttt{Locate} node of the phrase ``half sandwich'' outputs higher probabilities for both halves of sandwiches compared to other objects.
Prepositional phrases (``on right side'',``nearest'') correspond to \texttt{Relate} nodes in the computation graph.
\texttt{Relate} nodes calculate how likely objects are related to the grounding of objects with given prepositional phrase.
For instance, in Figure~\ref{fig1:c}, the \texttt{Relate} node of  ``nearest'' computes how likely the objects are related to the grounding of ``coffee mug'' with the relation ``nearest''.
We convert the phrases coming from branches in the parse tree to \texttt{Intersect} nodes.
It simply intersects two sets of groundings so that objects that have high likelihood in both branches will have high probabilities for the output (see the root node in Figure~\ref{fig1:d}). 
Since each node of this computation graph outputs a grounding for its subgraph, GroundNet is interpretable as a whole.
At each node, we can visualize how model's multiple predictions for objects propagates through the computation graph.

In following sections, we detail how we generate the computation graph and the neural modules used in GroundNet.

\subsection{Generating a Computation Graph}\label{ssec:fcg}

GroundNet processes the referring expression with a computation graph (Figure~\ref{fig1:c}) based on to the parse tree (Figure~\ref{fig1:b}) of the referring expression.
First, we parse the referring expression with Stanford Parser \cite{manning2014stanford}.
Then, we generate the computation graph (see Figure~\ref{fig1:b},~\ref{fig1:c} for an example) for a parse tree with a recursive algorithm (see Algorithm~\ref{alg:gct}).

\begin{algorithm}[H]
\caption*{\textbf{Algorithm 1:} Generate Computation Graph}
\begin{algorithmic}[1]
\Procedure{\textit{GenerateComputationGraph}}{tree}\label{alg:gct}
\State left\_NP  = \textit{FindNP}(tree.left) \label{alg:l1}
\State right\_NP = \textit{FindNP}(tree.right) \label{alg:l2}

\If{left\_NP == ""}
	\State return (\texttt{Locate} tree.text) \label{alg:l3}
\EndIf
\State  \texttt{Relate} = \textit{FindPP}(tree, [left\_NP, right\_NP])\label{alg:l4}
\State
\State left\_cg  = \textit{GenerateComputationGraph}(left\_NP) \label{alg:l5}
\State right\_cg = \textit{GenerateComputationGraph}(right\_NP) \label{alg:l6}
\State return (\texttt{Intersect} (left\_cg) (\texttt{Relate} right\_cg)) \label{alg:l7}
\EndProcedure
\end{algorithmic}
\end{algorithm}
Above, the function \textit{FindNP} finds the noun-phrase with the largest word span of given root node for left and right branches (line ~\ref{alg:l1},~\ref{alg:l2}).
If the tree does not have an NP subtree, it returns a \texttt{Locate} node (line ~\ref{alg:l3}). 

\textit{FindPP} extracts the words between noun-phrases to model the relationship between them and returns a \texttt{Relate} node (line ~\ref{alg:l4}).
For both left and right branches of the parse tree, the same algorithm is recursively called (lines ~\ref{alg:l5},~\ref{alg:l6}).
Finally, the sub-computation graphs of left and right branches are merged (line ~\ref{alg:l7}) into an \texttt{Intersect} node.

Each node in the computation graph is decorated with the phrase $T$ using the text span, i.e. constituents, of the corresponding parse tree node.
We filter out the function words such as determiners `a` and `the`.
For instance, the \texttt{Locate} on the left in Figure~\ref{fig1:c} has the span of words ``half sandwich'' from the corresponding noun phrase ``the half of a sandwich'' in Figure~\ref{fig1:b}.

In the following section, we explain the set of neural modules that we design for performing the localization of the referring expression on a composed computation graph.

\subsection{Neural Modules}\label{ssec:nnm}
We operationalize the computational graph for a referring expression into an end-to-end neural architecture by designing neural modules that represent each node of our graph.
First, let us introduce the notation for referring expression task.
For each referring expression, $(I,R,X)$ are inputs where $I$ is an image, $R$ is the set of bounding boxes $r_i$ of objects present in the image $I$, and  $X$ is a referring expression disambiguating a target object in bounding box $r^{*}$.
Our aim is to predict $r^{*}$ processing the referring expression in a computational graph with neural modules.
In addition to $(I,R,X)$, neural modules use the output of other neural modules and the text span $T$ of the computation node. 

We detail parameterization of neural modules in following subsections and visualize them in Figure~\ref{fig2} for clarity.
\figureten
\subsubsection{\texttt{Attend}} \label{sssec:am}
This module induces a text representation for \texttt{Locate} and \texttt{Relate} nodes. 
It takes the words ${\{w_i\}}_{i=1}^{\mid T \mid}$ and embeds them to a word vector ${\{e_i\}}_{i=1}^{\mid T \mid}$.
A 2-layer bidirectional LSTM network \cite{schuster1997bidirectional} processes embedded words.
Both forward and backward layer representations are concatenated for both layers into a single hidden representation for each word as follows:
\begin{align}
	h_i = [h_i^{(1,fw)} h_i^{(1,bw)} h_i^{(2,fw)} h_i^{(2,bw)}] \label{eq-am1}
\end{align}
The attention weights are computed with a linear projection using $W^{a}$:
\begin{align}
	a_i = \frac{exp( W^{a} h_i )}{\sum_{i=1}^{|T|}exp( W^{a} h_i )} \label{eq-am2}
\end{align}
The output of \texttt{Attend} is the weighted average of word vectors $e_i$ where the weights are attentions $a_i$.
\begin{align}
	f_{a}(T; \Theta_{a}) = \sum_{i=1}^{|T|} a_i e_i \label{eq-am3}
\end{align}
The learned parameters $\Theta_{a}$ of this module are the parameters of 2-layer bidirectional LSTM and scoring matrix $W^{a}$.
\subsubsection{\texttt{Locate}}\label{sssec:lm}
This module predicts which object is referred to for a text span, i.e. noun phrase, in the referring expression.
It computes the probability distribution over bounding boxes using the output of \texttt{Attend} and feature representations of bounding boxes.
For instance in Figure~\ref{fig1:c}, \texttt{Locate} node with input ``half sandwich''  localizes objects by scoring each bounding box.
\texttt{Locate} node does so by scoring how well the text span ``half sandwich'' matches the content of each bounding box. 

To represent a bounding box $r$, we use spatial and visual features.
First, visual features $r_{vis}$ for the bounding box are extracted using a convolutional neural network \cite{ren2015faster}.
Second, spatial features represent position and size of the bounding box.
We have 5-dimensional vectors for spatial features $r_{spat} = [\frac{x_{min}}{W_I}, \frac{y_{min}}{H_I}, \frac{x_{max}}{W_I}, \frac{y_{max}}{H_I}, \frac{S_r}{S_I}  ]$ where $S_r$ is the size and $[x_{min}, y_{min}, x_{max}, y_{max}]$ are coordinates for bounding box $r$ and $S_I$, $W_I$, $H_I$ are area, width, and the height of the input image $I$.
These two representations are concatenated as $r_{vis,spat} = [r_{vis} r_{spat}]$ for a bounding box $r$.

We follow the previous work \cite{hu2017modeling} for parametrization of \texttt{Locate}.
\begin{align}
	\hat{r}_{vis,spat} & = W_{vis,spat}^{loc} r_{vis,spat} \label{eq-lm1} \\ 
	z_{loc} & = \hat{r}_{vis,spat} \odot f_{a}(T) \label{eq-lm2} \\
    \hat{z}_{loc} & =  z_{loc} / \mid\mid z_{loc} \mid\mid_{2} \label{eq-lm3} \\ 
    s_{loc} & =  W^{loc}_{score} \hat{z}_{loc}  \label{eq-lm4} \\
    p_{loc} & =  softmax(s_{loc}) \label{eq-lm5} \\
    f_{loc}(T,R; \Theta_{loc}) & = p_{loc} \label{eq-lm6}
\end{align}
First, $r_{vis,spat}$ is projected to the same dimension as the text representation coming from the \texttt{Attend} (Eq~\ref{eq-lm1}).
Text and box representations are element-wise multiplied to get $z_{loc}$ for a joint representation of the text and bounding box. We normalize with L2-norm into $\hat{z}_{loc}$ (Eq~\ref{eq-lm2},~\ref{eq-lm3}).
Localization score $s_{loc}$ is calculated with a linear projection of the joint representation  (Eq~\ref{eq-lm4}).
Localization scores are fed to softmax to form a probability distribution $p_{loc}$ over boxes.
The learned parameters $\Theta_{loc}$ of this module are the  matrices $W_{vis,spat}^{loc}$ and $W^{loc}_{score}$.

\subsubsection{\texttt{Relate}} \label{sssec:rm} predicts how likely an object \textit{relates} to the other objects with some relation described by the node's text span.
For instance, the relation ``nearest'' in Figure~\ref{fig1:d} holds for half-sandwich pairs, and a half-sandwich and coffee mug pair.
Since the incoming \texttt{Locate} node to \texttt{Relate} outputs a high probability for the coffee mug, only objects near to coffee mug have a high probability.
GroundNet does so by first computing a relationship score matrix for boxes and multiplying the scoring matrix with the grounding input. 
We do not define a set of relationships for \texttt{Relate}, instead, model learns how objects relate to each other using module's text representation.
Specifically, this module computes a relationship score matrix $S_{rel}$ of size $R \times R$ consisting of scores for box $i$ and $j$ as follows:
\begin{align}
	\hat{r}_{i,j} & = W^{rel}_{spat} r_{i,j} \label{eq-rm1} \\
    z_{rel} & = \hat{r}_{i,j} \odot f_{a}(T) \label{eq-rm2} \\
    \hat{z}_{loc} & = z_{rel} / \mid\mid z_{rel} \mid\mid_{2} \label{eq-rm3} \\
    S_{rel}[i,j] & = W^{rel}_{score} \hat{z}_{rel} \label{eq-rm4} \\
    p_{rel} & = S_{rel} p \label{eq-rm5} \\
    f_{rel}(T,R,p; \Theta_{rel}) & = p_{rel} \label{eq-rm6}    
\end{align}
Above, spatial representations of boxes are concatenated as $r_{i,j} = [r_{i,spat}, r_{j,spat}]$ and projected into the same dimension as text representation (Eq~\ref{eq-rm1}). Similar to \texttt{Locate}, text and box representations are fused with element-wise multiplication and L2-normalization (Eq~\ref{eq-rm2},~\ref{eq-rm3}), then box pair is scored linearly (Eq~\ref{eq-rm4}). 

Finally, the probability distribution $p_{rel}$ over bounding boxes is calculated as $p_{rel} = S_{rel} p_{loc}$.
The learned parameters $\Theta_{rel}$ of this module are the  matrices $W^{rel}_{spat}$ and $W^{rel}_{score}$.

\subsubsection{\texttt{Intersect}}\label{sssec:combm}
This module combines groundings coming from two branches of the computation graph by simply multiplying object probabilities and normalizing it to form a probability distribution.
In the following section, we explain our experimental setup.
\section{Experiments}\label{sec:experiments}
Now, we detail our experimental setup.
In our experiments, we are interested in following research questions:
\begin{itemize}
\item \textbf{(RQ1)} How successful models are incorporating the syntax and how important the dynamic and modular computation in exploiting the syntactic information?
\item \textbf{(RQ2)} What are the accuracies of models for supporting objects and how these accuracies change depending on the syntactic information?
\end{itemize}
Now, we explain datasets used for our experiments.
\paragraph{Referring Expression Dataset.} We use the standard Google-Ref \cite{mao2016generation} benchmark for our experiments.
Google-Ref is a dataset consisting of around 26K images with 104K annotations.
We use "Ground-Truth" evaluation setting where the ground truth bounding box annotations from MSCOCO \cite{lin2014microsoft} are used.
\paragraph{Supporting Objects Dataset.} We also investigate the performances of models in terms of interpretability.
We measure the interpretability of a model by its accuracy on both target and supporting objects.
To this end, we present a new set of annotations on Google-Ref dataset.
First, we run a pilot study on MTurk where all bounding boxes and the referring expression present to annotators\footnote{We did not provide the parse trees to not bias the annotators.}.
Our in-house annotator has an agreement of 0.75 - a standard metric in word alignment literature \cite{graca2008building,ozdowska2008cross} with three turkers on a small validation set of 50 instances.
Overall, our annotator labeled 2400 instances -- but only 1023 had at least one supporting object bounding box.

\begin{table}[h]
\centering
\small
\begin{tabular}{@{}lrrrrr@{}}
\toprule
Number of Supporting Objects & 0    & 1   & 2   & 3  & 4 \\ \midrule
Number of Instances          & 1377 & 891 & 118 & 11 & 3 \\ \bottomrule
\end{tabular}
\caption{Statistics for the number of supporting objects for annotated 2400 instances.}\label{tab:stats}
\end{table}

We remind that the training data does not have any annotations for supporting objects.
Models should be able to predict supporting objects using only target object supervision and text input.
We should emphasize that our work is the first to report quantitative results on supporting object for the referring expression task and we release our annotation for future studies. Next, we provide details of our implementation.
\paragraph{Implementation Details.} We trained GroundNet with backpropagation.
We used stochastic gradient descent for 6 epochs with and initial learning rate of 0.01 and multiplied by 0.4 after each epoch.
Word embeddings were initialized with GloVe \cite{pennington2014glove} and finetuned during training.
We extracted features for bounding boxes using fc7 layer output of Faster-RCNN VGG-16 network \cite{ren2015faster} pre-trained on MSCOCO dataset \cite{lin2014microsoft}.
Hidden layer size of LSTM networks was searched over the range of \{64,128,...,1024\} and picked based on best validation split which is 2,5\% of training data separated from training split.
Following the previous work \cite{hu2017modeling}, we used official validation split as the test.
We initialized all parameters of the model with Xavier initialization \cite{glorot2010understanding} and used weight decay rate of 0.0005 as regularization.
Next, we explain models used in our experiments.
\tableseven
\paragraph{Baseline Models.}
We compare GroundNet to the recent models from the literature.
\textbf{RecursiveNN} \citet{socher2014grounded} use the recursive structure of syntactic parses of sentences to retrieve images described by the input sentence.
The text representation of a referring expression is recursively calculated following the parse tree of the referring expression.
The text representation at root node is jointly scored with bounding box representations and the highest scoring box is predicted.
\textbf{LSTM + CNN - MMI} \citet{mao2016generation} use LSTMs for processing the referring expression and CNNs for extracting features for bounding boxes and the whole image.
Model is trained with Maximum Mutual Information training. 
\textbf{LSTM + CNN - MMI+visdif} \citet{yu2016modeling} introduce contextual features for a bounding box by calculating differences between visual features for object pairs. 
\textbf{LSTM + CNN - MIL}\footnote{Originally the authors use a new test split, whereas, we report results for the standard split of the dataset for this model.} \citet{nagaraja16refexp} score object-supporting object pairs. The pair with the highest score is predicted.
They use Multi Instance Learning for training the model.
\textbf{CMN}\footnote{We report results for our reimplementation of this model where we did hyperparameter search the same as our model.} \citet{hu2017modeling} introduce a neural module network with a tuple of object-relationship-subject nodes.
The text representation of tuples are calculated with an attention mechanism \cite{bahdanau2014neural} over the referring expression.
We also report results for \textbf{CMN - syntax guided} when a parse tree is used for extracting the object-relationship-subject tuples.
\paragraph{GroundNet with varying level of syntax.} We investigate the effect of the syntax varying the level of use of the syntactic structure for GroundNet.
\textbf{GroundNet} is the original model presented in the previous section where each node in computation graph uses the node's text span for \texttt{Attend}. For
\textbf{GroundNet-syntax-guided \texttt{Locate}} model, \texttt{Locate} nodes use the node's text span as an input to the \texttt{Attend} module. Whereas for \texttt{Relate} nodes can use all referring expression for inducing the text representation.
For \textbf{GroundNet-free-form} model, Both \texttt{Locate} and \texttt{Relate} nodes use all of the referring expression as the input to \texttt{Attend}.
Next, we explain our evaluation metrics used in our experiments. 
\paragraph{Evaluation.}
To evaluate models for referring expression task we use the standard metric of accuracy.
For evaluation of supporting objects, when there are multiple supporting objects, we consider a supporting object prediction as accurate only if at least one supporting object is correctly classified.
To evaluate approaches modeling the supporting objects we use following methods.
For LSTN+CNN-MIL, we use the context object of the maximum scoring target-context object pair as the supporting object.
For CMN, we use the object with the maximum object  score of a subject-relation-object tuple as the prediction for the supporting object.
For GroundNet, we use the object with maximum probability as a prediction for intermediate nodes in the computation graph.
In the following section, we discuss results of our experiments.
\figurecpick
\section{Results}\label{sec:results}
We presented overall results in Table~\ref{tab1} for the compared models.
We now discuss columns of the Table~\ref{tab1}.
\paragraph{(RQ1) Syntax, Dynamic Computation, and Modularity.} GroundNet variations achieve the best results among syntax-based models.
``Recursive NN'' homogeneously processes the referring expression throughout the parse tree structure.
On the other hand, GroundNet modularly parameterizes multi-modal processing of localization and relationships.
``CMN - syntax guided'' has a fixed computation graph of a subject-relation-object tuple, whereas, GroundNet has a dynamic computation graph for each instance, thus, a varying number of computation nodes are induced.
When compared to other syntax-based approaches, GroundNet results show that a dynamic \textit{and} modular architecture is essential to achieve competitive results with a syntax-based approach.
\paragraph{(RQ2) Syntax for Supporting Objects.}
Our model achieves the highest accuracy on localizing the supporting objects when its modules are guided by syntax.
``LSTM+CNN-MIL'' and CMN does not exploit the syntax of the referring expression and poorly performs in localizing supporting objects.
When we relax the syntactic guidance of GroundNet by letting all modules to attend to all of the referring expression, ``GroundNet-free-form'' also performs poorly on localizing supporting objects.
These results suggest that leveraging syntax is essential in localizing supporting objects and there might be a tradeoff between being interpretable and being accurate for models.
We qualitatively show a couple of instances from test set GroundNet and CMN in Figure~\ref{fig3}.
As an example, for the first instance, both GroundNet and CMN successfully predict the target object.
GroundNet is able to localize both supporting objects (i.e. the girl and the disc) mentioned in the referring expression, whereas, CMN fails to localize the supporting objects.
Next, we review the previous work related to GroundNet.
\section{Related Work}\label{sec:related}
Referring expressiong recognition is a well-studied problem in human-robot interaction \cite{chai2004probabilistic,zender2009situated,tellex2011approaching,lemaignan2011you,fang2012integrating,williams2016situated}. Here, we focus on more closely related studies where visual context is a rich set of real-world images or language with rich vocabulary is modeled with compositionality.
\paragraph{Grounding Referential Expressions.}
The most of the recent work \cite{mao2016generation,hu2016natural,rohrbach2016grounding,fukui2016,yu2016modeling,nagaraja16refexp} addresses grounding referential expression task  with a fixed computation graph.
In earlier studies \cite{mao2016generation,hu2016natural,rohrbach2016grounding,fukui2016}, the bounding boxes are scored based on their CNN and spatial features along with features for the whole image.
Since each box is scored in isolation, these methods ignore the object relationships.
More recent studies \cite{yu2016modeling,nagaraja16refexp,hu2017modeling} show that modeling relationship between objects improves the accuracy of models.
GroundNet has a dynamic computation graph and models the relationship between objects.
\paragraph{Modular Neural Architectures.} Neural Module Networks (NMN) \cite{andreas2016neural,andreas2016learning} is a general framework for modeling compositionality of language using neural modules.
A computation graph with neural modules as nodes is generated based on a parse tree of the input text.
GroundNet shares the principles of this framework.
We design GroundNet for referring expression task  restricting each node grounded in the input image which keeps network interpretable throughout the computation.

Compositional Modular Networks (CMN) \cite{hu2017modeling} is also an instant of NMN aiming to remove language parser from the generation of computation graph by inducing text representations to localization and relationship modules using an attention mechanism.
Their computation graph is fixed to the subject-relation-subject tuple but the input is dynamically constructed for modules.
Our model, on the other hand, can handle multiple relationships mentioned in referring expressions (see the first row of Figure~\ref{fig3}).
We should note that CMN is a special case of GroundNet where the syntax is fixed to a triplet of $Locate_{subject},Relate, Locate_{obj}$ and each node composes a text representation with whole referring expression.
\paragraph{Syntax with Vision.}
Similar to our work, \citet{gorniak2004grounded} study a syntax-based approach for grounding referring expressions.
However, they use a synthetic visual scene of identical shapes with varying colors and a synthetic grammar for language.
\citet{golland2010game} introduce a game-theoretic model successfully leverages syntax for grounding reference expressions for synthetic scenes.
\cite{matuszek2012joint} presents a semantic parsing model with Combinatory Category Grammar for referring expression recognition that jointly learns grounding of objects and their attributes. The model is able to induce latent logical forms when bootstrapped with a supervised training stage.
\citet{berzak2015do} use visual context to address linguistic ambiguities. Similarly,
\citet{gordon2016} use the visual context for solving prepositional phrase attachment resolution (PPAR) for sentences describing a scene.
Unlike our model, their model relies on multiple parse trees and multiple segmentations of an image coming from a black-box image segmenter.
Our model can also be extended to address PPAR setting where
we only need to ground-truth object annotations for roots of multiple parse trees for the input sentences.
\citet{wang2016structured} introduce a model localizing phrases in sentences that describe an image.
However, their model relies on the annotation of phrase-object pairs.   
GroundNet only uses target object annotations and there is no supervision for supporting objects.
\citet{xiao2017weakly} aim to address localization of phrases on region masks.
Similar to our approach, they do not rely on ground-truth masks during training.
However, unlike GroundNet, their model does not model relationship between objects.
\section{Conclusion}\label{sec:conclusion}
In this work, we present GroundNet, a compositional neural module network designed for the task of grounding referring expressions.
We also introduce a novel auxiliary task and an annotation for localizing the supporting objects.

Our experiments on a standard benchmark show that GroundNet is the first model that successfully incorporates syntactic information for the referring expression task.
This syntactic information helps GroundNet achieve state-of-the-art results in localizing supporting objects. 
Our results show that recent models are unsuccessful at localizing supporting objects.
This suggests that current solutions to referring expression task come with an interpretability-accuracy trade-off.
Our approach substantially improves supporting object localization, while maintaining high accuracy, thus representing a new and more desirable point along the trade-off trajectory.

We believe future work might extend our work with following insights.
First, while generating the computation graph GroundNet, we drop the determiners.
However, the indefiniteness of a noun could be helpful in localizing an object.
Second, GroundNet processes the computation graph in a bottom-up fashion.
An approach combining the sequential processing of the referring expression with the bottom-up structural processing of GroundNet could model expectation-driven effects of language which may result in more accurate grounding throughout the computation graph.

\section*{Acknowledgments}
This project was partially supported by Oculus
and Yahoo InMind research grants. The authors would like to thank anonymous reviewers and the members of MultiComp Lab at CMU for their valuable feedback.
\bibliographystyle{aaai}
\bibliography{refs}

\begin{thebibliography}{}

\bibitem[\protect\citeauthoryear{Andreas \bgroup et al\mbox.\egroup
  }{2016a}]{andreas2016learning}
Andreas, J.; Rohrbach, M.; Darrell, T.; and Klein, D.
\newblock 2016a.
\newblock Learning to compose neural networks for question answering.
\newblock In {\em Proceedings of the 2016 Conference of the North American
  Chapter of the Association for Computational Linguistics: Human Language
  Technologies},  1545--1554.
\newblock San Diego, California: Association for Computational Linguistics.

\bibitem[\protect\citeauthoryear{Andreas \bgroup et al\mbox.\egroup
  }{2016b}]{andreas2016neural}
Andreas, J.; Rohrbach, M.; Darrell, T.; and Klein, D.
\newblock 2016b.
\newblock Neural module networks.
\newblock In {\em Proceedings of the IEEE Conference on Computer Vision and
  Pattern Recognition (CVPR)},  39--48.

\bibitem[\protect\citeauthoryear{Bahdanau, Cho, and
  Bengio}{2014}]{bahdanau2014neural}
Bahdanau, D.; Cho, K.; and Bengio, Y.
\newblock 2014.
\newblock Neural machine translation by jointly learning to align and
  translate.
\newblock {\em arXiv preprint arXiv:1409.0473}.

\bibitem[\protect\citeauthoryear{Berzak \bgroup et al\mbox.\egroup
  }{2015}]{berzak2015do}
Berzak, Y.; Barbu, A.; Harari, D.; Katz, B.; and Ullman, S.
\newblock 2015.
\newblock Do you see what i mean? visual resolution of linguistic ambiguities.
\newblock In {\em Proceedings of the 2015 Conference on Empirical Methods in
  Natural Language Processing},  1477--1487.
\newblock Association for Computational Linguistics.

\bibitem[\protect\citeauthoryear{Chai, Hong, and
  Zhou}{2004}]{chai2004probabilistic}
Chai, J.~Y.; Hong, P.; and Zhou, M.~X.
\newblock 2004.
\newblock A probabilistic approach to reference resolution in multimodal user
  interfaces.
\newblock In {\em Proceedings of the 9th international conference on
  Intelligent user interfaces},  70--77.
\newblock ACM.

\bibitem[\protect\citeauthoryear{Christie \bgroup et al\mbox.\egroup
  }{2016}]{gordon2016}
Christie, G.; Laddha, A.; Agrawal, A.; Antol, S.; Goyal, Y.; Kochersberger, K.;
  and Batra, D.
\newblock 2016.
\newblock Resolving language and vision ambiguities together: Joint
  segmentation {\&} prepositional attachment resolution in captioned scenes.
\newblock In {\em Proceedings of the 2016 Conference on Empirical Methods in
  Natural Language Processing},  1493--1503.
\newblock Association for Computational Linguistics.

\bibitem[\protect\citeauthoryear{Fang, Liu, and
  Chai}{2012}]{fang2012integrating}
Fang, R.; Liu, C.; and Chai, J.~Y.
\newblock 2012.
\newblock Integrating word acquisition and referential grounding towards
  physical world interaction.
\newblock In {\em Proceedings of the 14th ACM international conference on
  Multimodal interaction},  109--116.
\newblock ACM.

\bibitem[\protect\citeauthoryear{Fukui \bgroup et al\mbox.\egroup
  }{2016}]{fukui2016}
Fukui, A.; Park, D.~H.; Yang, D.; Rohrbach, A.; Darrell, T.; and Rohrbach, M.
\newblock 2016.
\newblock Multimodal compact bilinear pooling for visual question answering and
  visual grounding.
\newblock In {\em Proceedings of the 2016 Conference on Empirical Methods in
  Natural Language Processing},  457--468.
\newblock Austin, Texas: Association for Computational Linguistics.

\bibitem[\protect\citeauthoryear{Glorot and
  Bengio}{2010}]{glorot2010understanding}
Glorot, X., and Bengio, Y.
\newblock 2010.
\newblock Understanding the difficulty of training deep feedforward neural
  networks.
\newblock In {\em Aistats}, volume~9,  249--256.

\bibitem[\protect\citeauthoryear{Golland, Liang, and
  Klein}{2010}]{golland2010game}
Golland, D.; Liang, P.; and Klein, D.
\newblock 2010.
\newblock A game-theoretic approach to generating spatial descriptions.
\newblock In {\em Proceedings of the 2010 conference on empirical methods in
  natural language processing},  410--419.
\newblock Association for Computational Linguistics.

\bibitem[\protect\citeauthoryear{Gorniak and Roy}{2004}]{gorniak2004grounded}
Gorniak, P., and Roy, D.
\newblock 2004.
\newblock Grounded semantic composition for visual scenes.
\newblock {\em Journal of Artificial Intelligence Research} 21:429--470.

\bibitem[\protect\citeauthoryear{Graca \bgroup et al\mbox.\egroup
  }{2008}]{graca2008building}
Graca, J.; Pardal, J.~P.; Coheur, L.; and Caseiro, D.
\newblock 2008.
\newblock Building a golden collection of parallel multi-language word
  alignment.
\newblock In {\em LREC}.

\bibitem[\protect\citeauthoryear{Grice}{1975}]{grice1975logic}
Grice, H.~P.
\newblock 1975.
\newblock Logic and conversation.
\newblock {\em 1975}  41--58.

\bibitem[\protect\citeauthoryear{Hu \bgroup et al\mbox.\egroup
  }{2016}]{hu2016natural}
Hu, R.; Xu, H.; Rohrbach, M.; Feng, J.; Saenko, K.; and Darrell, T.
\newblock 2016.
\newblock Natural language object retrieval.
\newblock In {\em Proceedings of the IEEE Conference on Computer Vision and
  Pattern Recognition},  4555--4564.

\bibitem[\protect\citeauthoryear{Hu \bgroup et al\mbox.\egroup
  }{2017}]{hu2017modeling}
Hu, R.; Rohrbach, M.; Andreas, J.; Darrell, T.; and Saenko, K.
\newblock 2017.
\newblock Modeling relationships in referential expressions with compositional
  modular networks.

\bibitem[\protect\citeauthoryear{Lemaignan \bgroup et al\mbox.\egroup
  }{2011}]{lemaignan2011you}
Lemaignan, S.; Ros, R.; Alami, R.; and Beetz, M.
\newblock 2011.
\newblock What are you talking about? grounding dialogue in a perspective-aware
  robotic architecture.
\newblock In {\em RO-MAN, 2011 IEEE},  107--112.
\newblock IEEE.

\bibitem[\protect\citeauthoryear{Lin \bgroup et al\mbox.\egroup
  }{2014}]{lin2014microsoft}
Lin, T.-Y.; Maire, M.; Belongie, S.; Hays, J.; Perona, P.; Ramanan, D.;
  Doll{\'a}r, P.; and Zitnick, C.~L.
\newblock 2014.
\newblock Microsoft coco: Common objects in context.
\newblock In {\em European Conference on Computer Vision},  740--755.
\newblock Springer.

\bibitem[\protect\citeauthoryear{Manning \bgroup et al\mbox.\egroup
  }{2014}]{manning2014stanford}
Manning, C.~D.; Surdeanu, M.; Bauer, J.; Finkel, J.~R.; Bethard, S.; and
  McClosky, D.
\newblock 2014.
\newblock The stanford corenlp natural language processing toolkit.
\newblock In {\em ACL (System Demonstrations)},  55--60.

\bibitem[\protect\citeauthoryear{Mao \bgroup et al\mbox.\egroup
  }{2016}]{mao2016generation}
Mao, J.; Huang, J.; Toshev, A.; Camburu, O.; Yuille, A.~L.; and Murphy, K.
\newblock 2016.
\newblock Generation and comprehension of unambiguous object descriptions.
\newblock In {\em Proceedings of the IEEE Conference on Computer Vision and
  Pattern Recognition (CVPR)},  11--20.

\bibitem[\protect\citeauthoryear{Matuszek* \bgroup et al\mbox.\egroup
  }{2012}]{matuszek2012joint}
Matuszek*, C.; FitzGerald*, N.; Zettlemoyer, L.; Bo, L.; and Fox, D.
\newblock 2012.
\newblock {A Joint Model of Language and Perception for Grounded Attribute
  Learning}.
\newblock In {\em Proc.~of the 2012 International Conference on Machine
  Learning}.

\bibitem[\protect\citeauthoryear{Nagaraja, Morariu, and
  Davis}{2016}]{nagaraja16refexp}
Nagaraja, V.; Morariu, V.; and Davis, L.
\newblock 2016.
\newblock Modeling context between objects for referring expression
  understanding.
\newblock In {\em ECCV}.

\bibitem[\protect\citeauthoryear{Ozdowska}{2008}]{ozdowska2008cross}
Ozdowska, S.
\newblock 2008.
\newblock Cross-corpus evaluation of word alignment.
\newblock In {\em LREC}.

\bibitem[\protect\citeauthoryear{Pennington, Socher, and
  Manning}{2014}]{pennington2014glove}
Pennington, J.; Socher, R.; and Manning, C.~D.
\newblock 2014.
\newblock Glove: Global vectors for word representation.
\newblock In {\em EMNLP}, volume~14,  1532--1543.

\bibitem[\protect\citeauthoryear{Ren \bgroup et al\mbox.\egroup
  }{2015}]{ren2015faster}
Ren, S.; He, K.; Girshick, R.; and Sun, J.
\newblock 2015.
\newblock Faster r-cnn: Towards real-time object detection with region proposal
  networks.
\newblock In {\em Advances in neural information processing systems},  91--99.

\bibitem[\protect\citeauthoryear{Rohrbach \bgroup et al\mbox.\egroup
  }{2016}]{rohrbach2016grounding}
Rohrbach, A.; Rohrbach, M.; Hu, R.; Darrell, T.; and Schiele, B.
\newblock 2016.
\newblock Grounding of textual phrases in images by reconstruction.
\newblock In {\em European Conference on Computer Vision},  817--834.
\newblock Springer.

\bibitem[\protect\citeauthoryear{Schuster and
  Paliwal}{1997}]{schuster1997bidirectional}
Schuster, M., and Paliwal, K.~K.
\newblock 1997.
\newblock Bidirectional recurrent neural networks.
\newblock {\em IEEE Transactions on Signal Processing} 45(11):2673--2681.

\bibitem[\protect\citeauthoryear{Socher \bgroup et al\mbox.\egroup
  }{2014}]{socher2014grounded}
Socher, R.; Karpathy, A.; Le, Q.~V.; Manning, C.~D.; and Ng, A.~Y.
\newblock 2014.
\newblock Grounded compositional semantics for finding and describing images
  with sentences.
\newblock {\em Transactions of the Association for Computational Linguistics}
  2:207--218.

\bibitem[\protect\citeauthoryear{Tellex \bgroup et al\mbox.\egroup
  }{2011}]{tellex2011approaching}
Tellex, S.; Kollar, T.; Dickerson, S.; Walter, M.~R.; Banerjee, A.~G.; Teller,
  S.; and Roy, N.
\newblock 2011.
\newblock Approaching the symbol grounding problem with probabilistic graphical
  models.
\newblock {\em AI magazine} 32(4):64--76.

\bibitem[\protect\citeauthoryear{Wang \bgroup et al\mbox.\egroup
  }{2016}]{wang2016structured}
Wang, M.; Azab, M.; Kojima, N.; Mihalcea, R.; and Deng, J.
\newblock 2016.
\newblock Structured matching for phrase localization.
\newblock In {\em European Conference on Computer Vision},  696--711.
\newblock Springer.

\bibitem[\protect\citeauthoryear{Williams \bgroup et al\mbox.\egroup
  }{2016}]{williams2016situated}
Williams, T.; Acharya, S.; Schreitter, S.; and Scheutz, M.
\newblock 2016.
\newblock Situated open world reference resolution for human-robot dialogue.
\newblock In {\em The Eleventh ACM/IEEE International Conference on Human Robot
  Interaction},  311--318.
\newblock IEEE Press.

\bibitem[\protect\citeauthoryear{Xiao, Sigal, and Lee}{2017}]{xiao2017weakly}
Xiao, F.; Sigal, L.; and Lee, Y.~J.
\newblock 2017.
\newblock Weakly-supervised visual grounding of phrases with linguistic
  structures.

\bibitem[\protect\citeauthoryear{Yu \bgroup et al\mbox.\egroup
  }{2016}]{yu2016modeling}
Yu, L.; Poirson, P.; Yang, S.; Berg, A.~C.; and Berg, T.~L.
\newblock 2016.
\newblock Modeling context in referring expressions.
\newblock In {\em European Conference on Computer Vision},  69--85.
\newblock Springer.

\bibitem[\protect\citeauthoryear{Zender, Kruijff, and
  Kruijff-Korbayov{\'a}}{2009}]{zender2009situated}
Zender, H.; Kruijff, G.-J.~M.; and Kruijff-Korbayov{\'a}, I.
\newblock 2009.
\newblock Situated resolution and generation of spatial referring expressions
  for robotic assistants.
\newblock In {\em IJCAI},  1604--1609.

\end{thebibliography}
\end{document}